\documentclass[runningheads]{llncs}
\usepackage[T1]{fontenc}

\usepackage[table,xcdraw]{xcolor}
\usepackage{graphicx}
\usepackage{algorithm}
\usepackage{algorithmic}
\usepackage{amsmath} 
\usepackage{amsfonts} 
\usepackage{amssymb} 
\usepackage{bm} 
\usepackage{colortbl} 
\usepackage[table]{xcolor} 
\usepackage{graphicx}
\usepackage{amsmath}
\usepackage{amssymb}
\usepackage{booktabs}
\usepackage{multirow}
\usepackage{booktabs}
\usepackage{algorithm}
\usepackage{algorithmic}
\usepackage{enumitem}
\definecolor{grayish}{RGB}{240, 240, 240} 
\definecolor{grayish_2}{RGB}{230, 230, 230} 
\usepackage{hyperref}
\usepackage{adjustbox}
\hypersetup{
    colorlinks=true,
    linkcolor=blue,
    citecolor=blue,
    urlcolor=blue
}
\begin{document}
%
\title{TiBiX: Leveraging Temporal Information for Bidirectional  X-ray and Report Generation}
\titlerunning{TiBiX}

\author{Santosh Sanjeev\inst{1},
Fadillah Adamsyah Maani\inst{1},
Arsen Abzhanov\inst{1}, Vijay Ram Papineni\inst{2}, Ibrahim Almakky\inst{1}, Bart\l omiej W. Papie\.z\inst{3} \and Mohammad Yaqub\inst{1}}
\authorrunning{S. Sanjeev et al.}
%
\institute{Mohamed bin Zayed University of Artificial Intelligence, Abu Dhabi, UAE \\
\email{\{firstname.lastname\}@mbzuai.ac.ae} \\ \and
Sheikh Shakhbout Medical City, Abu Dhabi, UAE \\ 
\email{\{vpapineni\}@ssmc.ae}
\and
 Big Data Institute, University of Oxford, UK \\
\email{\{firstname.lastname\}@bdi.ox.ac.uk}
}


%
%
%

%
%
\maketitle              
%
\begin{abstract}
With the emergence of vision language models in the medical imaging domain, numerous studies have focused on two dominant research activities: (1) report generation from Chest X-rays (CXR), and (2) synthetic scan generation from text or reports. Despite some research incorporating multi-view CXRs into the generative process, prior patient scans and reports have been generally disregarded. This can inadvertently lead to the leaving out of important medical information, thus affecting generation quality. To address this, we propose \textbf{TiBiX}: Leveraging \textbf{T}emporal \textbf{i}nformation for \textbf{Bi}directional \textbf{X}-ray and Report Generation. Considering previous scans, our approach facilitates bidirectional generation, primarily addressing two challenging problems: (1) generating the current image from the previous image and current report and (2) generating the current report based on both the previous and current images. Moreover, we extract and release a curated temporal benchmark dataset derived from the MIMIC-CXR dataset, which focuses on temporal data. Our comprehensive experiments and ablation studies explore the merits of incorporating prior CXRs and achieve state-of-the-art (SOTA) results on the report generation task. Furthermore, we attain on-par performance with SOTA image generation efforts, thus serving as a new baseline in longitudinal bidirectional CXR-to-report generation. The code is available at \url{https://github.com/BioMedIA-MBZUAI/TiBiX}.


\end{abstract}

\section{Introduction}



Image and text generation advancements have propelled the progress in Chest X-ray (CXR) and report generation tasks. The specific focus on CXR stems from the time-consuming and error-prone nature of the manual report composition task, putting a huge strain on radiologists \cite{Johnson2019Dec}. Moreover, the quality of manually composed reports depends on the radiologist's experience, which makes accurate report generation a good aid for inexperienced practitioners. To this effect, some research has been conducted in the domain of CXR image \cite{roentgen,weber2023cascaded,packhauser2023generation} and report \cite{miccai_report_1,miccai_report_3,miccai_report_4,miccai_report2} generation. 

Transformer-based models like those in \cite{r2gen,r2gencmn} have shown promising performance, especially with the introduction of expert tokens in \cite{metransformer}. Strategies such as integrating knowledge graphs \cite{yang2022knowledge,huang2023kiut}, addressing class-level bias \cite{ppked}, utilizing disease tags \cite{wang2022medical,wang2021self}, and applying contrastive learning \cite{contrastivereport} aim to enhance report quality. In parallel, advancements in natural image generation, driven by techniques like Generative Adversarial Networks (GANs) \cite{vqgan,attngan}, have sparked interest in generating CXR images from radiology reports aiming to improve a model's interpretability, helping build user trust through visual evidence \cite{Lanfredi2019Oct}. However, the translation of radiology reports into CXR images is fraught with challenges, mainly stemming from the linguistic complexities and varied structures of radiology reports, hindering accurate representation in image form \cite{weber2023cascaded,roentgen}. Despite efforts to confront these obstacles, existing methods often fail to capture the intricate relationships between textual descriptions and corresponding image features, resulting in suboptimal performance in CXR image generation. Moreover, these methods overlook the temporal aspect of patient data which is very important for medical diagnosis. 

There are few works exploring the temporal domain for generation tasks \cite{biovilt,miccai2023,controllable_temporal}. Unlike the previous works, which focus on only report generation, we focus on bidirectional CXR and report generation. Furthermore, most works require both the prior and current scan images, which is generally not the case in real-life scenarios as patients might not always have prior studies. Recent advancements in bidirectional approaches \cite{kim2022verse,huang2022vlg} have shown promise in bridging the gap between image-to-text and text-to-image generation tasks. In the medical domain, UniXGen was proposed for bidirectional CXR-to-report generation, leveraging multiple CXR views from the same patient during the same visit to enhance report accuracy \cite{unixgen}. Previous research has focused on generating reports or images at single time points, overlooking the longitudinal aspect of patient data \cite{Kayser2022Sep}. 

\begin{figure*}[t]
    \centering
    \includegraphics[width=\textwidth]{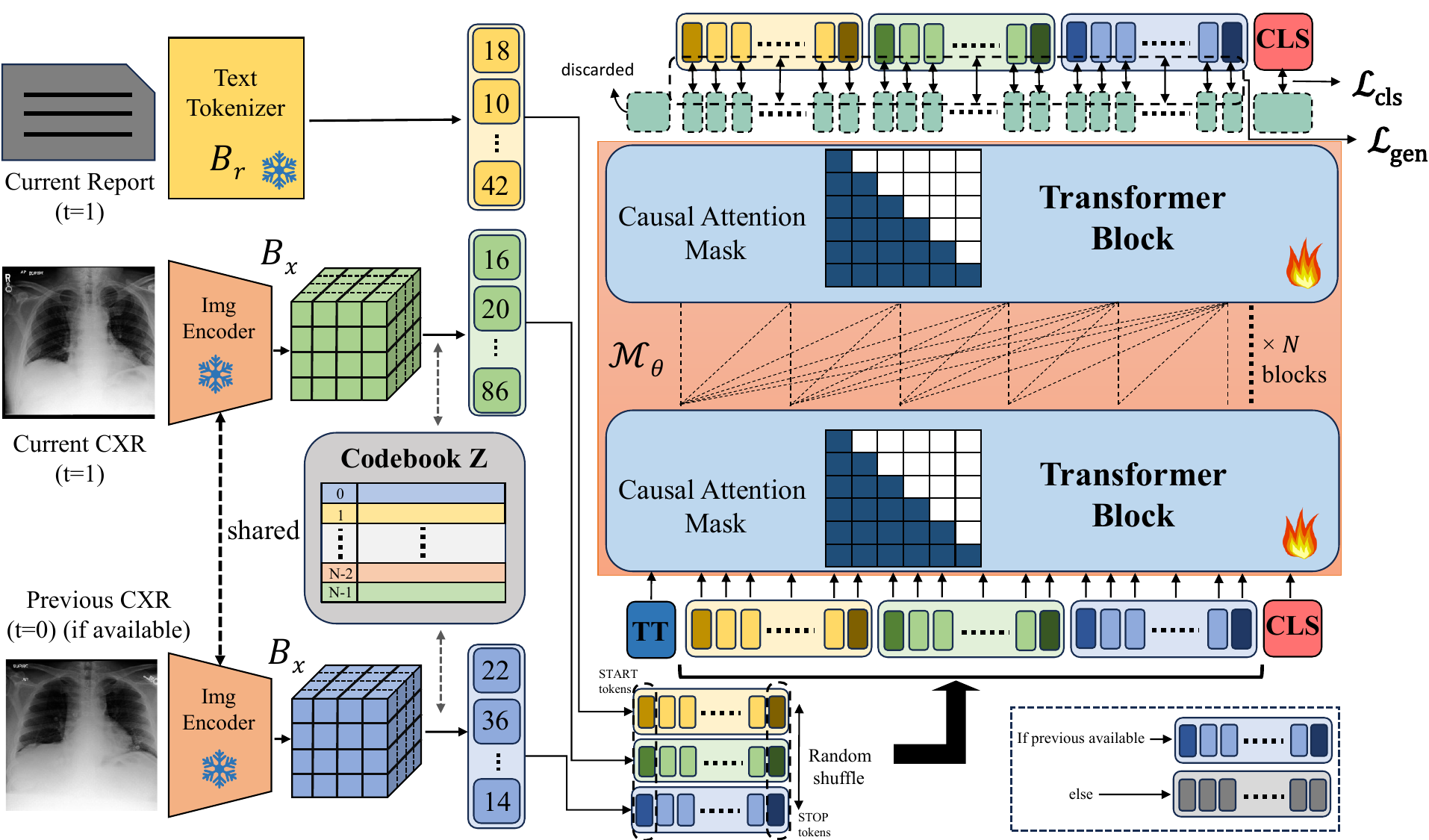} 
    \caption{Our temporal bidirectional CXR-to-report generation framework which can deal with three inputs (i.e. current report, current CXR, and previous CXR). We utilize an image tokenizer $B_x$ and text tokenizer $B_r$ to tokenize CXR image(s) and report, respectively. A transformer-based model $\mathcal{M}_\theta$ with \textit{causal} attention is implemented to handle the bidirectional generation task in the \textit{auto-regressive} manner. The input sequence of $\mathcal{M}_\theta$ consists of images and text tokens, a temporal token (\texttt{TT}) which encodes the time interval between two consecutive CXR scans, and a learnable \textit{cls} (class) token. We assign \texttt{TT} as the first input sequence and \textit{cls} as the last sequence. During training, the order of the current report, current CXR, and previous CXR are shuffled, while the missing modality is placed at the last part of the input sequence during inference.}
    \label{fig:methodology}
\end{figure*}

In this paper, we propose Chest X-ray and Report Bidirectional Generation along the temporal domain. Our main contributions are summarized as follows:
\begin{itemize}
    \item We propose a novel approach to integrate temporal data for the bidirectional generation task by designing a framework consisting of a transformer-based model equipped with causal attention layers that provide an effective way to learn temporal information while also reducing computational complexity. We also introduce a $cls$ (class) token guidance approach based on the patient's overall diagnosis and also a temporal token, considering the varying time intervals between patient visits.
    \item We achieve state-of-the-art (SOTA) performance on the benchmark dataset for the report generation task and achieve comparable performance to the (SOTA) on the image generation task. We also achieve promising results on the classification metrics of images and reports. Furthermore, we support the inclusion of temporal scans by conducting an extensive analysis.
    \item We extract and release the MIMIC-T dataset, curated from the MIMIC-CXR dataset and consists of longitudinal data of patients extracted from the MIMIC-CXR dataset.
\end{itemize}

To the best of our knowledge, this is the first effort to include prior(temporal) CXRs and reports for the bidirectional CXR-to-report generation task.

\section{Methodology}
\label{methodology}



We formulate the temporal bidirectional CXR-to-report generation task starting from a set of patients $P = \{p_1, p_2, \dots, p_n\}$. For each patient $p$, we have a set of CXR images $X_p = \{x_{(i)}\}_{i={t_0}}^{t_{m_p}}$ along with a set of radiology reports $R_p = \{r_{(i)}\}_{i={t_0}}^{t_{m_p}}$, where $m_p \ge 0$ is the total number of time steps and $r_{(t_j)}$ is the corresponding radiology report for $x_{(t_j)}$ captured at time step $t_{j}$. Therefore, the total number of samples available is $M = \sum_{p=1}^n{(m_p + 1)}$.
We define the labels for a patient $p$ at time step $t_j$ by $L_{(t_j)} = \{l_{(i)}\}_{i=1}^{c}$, where $l_{(i)} \in \{0,1\}$ and $c$ is the total number of possible pathologies.
We also define the variable time step $\delta = (t_j - t_{j-1})$ as the time interval between 2 consecutive studies, where $\delta > 0$. As such, the objective of the temporal bidirectional CXR-to-report generation task would be to estimate $r_{(t_j)}$ given $(x_{(t_j)}, x_{t_{(j-1)}})$, while also estimating $x_{(t_j)}$ given $(r_{(t_j)}, x_{t_{(j-1)}})$. To learn this, we employ a transformer-based model with causal attention, which is described later in this section, starting with the multi-modal input tokenization and ending with the auto-regressive training process. The high-level visualization of our framework is depicted in Fig. \ref{fig:methodology}.

\subsection{Multimodal Tokenization}

This work considers three modalities: current report, current CXR, and previous CXR. To facilitate a simple yet effective and memory-efficient multimodal interaction, we tokenize every modality accordingly. We use a CXR image tokenizer $B_x(x)=\mathcal{I}^x \in \mathbb{Z}^{N_x}$ and a report text tokenizer $B_r(r)=\mathcal{I}^r \in \mathbb{Z}^{N_r}$, where $N_x$ and $N_r$ are the number of image and text tokens. We adopt $B_x$ from VQGAN \cite{vqgan} tokenizer, which is composed of two primary components: an image encoder responsible for generating context-rich latent representations and a codebook used to map the representations to a quantized encoding. The main component of our bidirectional generation framework $\mathcal{M}_\theta$ is based on a transformer model. We, add unique \texttt{START} and \texttt{STOP} tokens for every modality, increasing the size of the image and text tokens to $\bar{\mathcal{I}}^x \in \mathbb{Z}^{N_x+2}$ and $\bar{\mathcal{I}}^r \in \mathbb{Z}^{N_r+2}$, respectively to avoid confusion between the modalities, as it would treat all input tokens uniformly. The time interval $\delta$ between two consecutive samples is also input to $\mathcal{M}_\theta$ as a temporal token ($\mathcal{I}^t$). This is added to further improve the model's prognosis capability by mimicking clinicians' approach to leveraging the time gap between different scans. Finally, we add a learnable token $cls \subset \theta$ to predict the pathologies existing in the patient's modalities $L_{(t_j)}$, allowing $\mathcal{M}_\theta$ to learn from the overall diagnosis. 

\subsection{Multimodal Generation Framework}
\noindent \textbf{Transformer-based Model:} $\mathcal{M}_\theta$ carries out the temporal bidirectional CXR-to-report generation task as follows: 
\begin{equation}
    \mathcal{M}_\theta(\mathcal{I}^t, \bar{\mathcal{I}}^{m_1}, \bar{\mathcal{I}}^{m_2}, \bar{\mathcal{I}}^{m_3}) = \hat{\mathcal{I}}^{m_1}, \hat{\mathcal{I}}^{m_2}, \hat{\mathcal{I}}^{m_3}, \hat{L}_t
\label{eq:tranformer_func}
\end{equation}
where $\hat{\mathcal{I}}^{m} \approx \bar{\mathcal{I}}^{m}$, $\hat{L}_t$ is the prediction of the patient's diagnosis, and $\{\bar{\mathcal{I}}^{m_1}, \bar{\mathcal{I}}^{m_2}, \bar{\mathcal{I}}^{m_3}\}$ are sampled from set $\{\bar{\mathcal{I}}^{x_{(t_{j})}}, \bar{\mathcal{I}}^{r_{(t_{j})}}, \bar{\mathcal{I}}^{x_{(t_{j-1})}}\}$ without replacement. Our main training goal is to learn parameters $\theta$ so that $\mathcal{M}_\theta$ is capable of performing Eq. \eqref{eq:tranformer_func} by optimizing the following objective:
\[
\min_{\theta}\frac{1}{M} \sum_{i=1}^{M}\Big(\mathcal{L}_{gen}\big((\hat{\mathcal{I}}^{m_1}_i, \hat{\mathcal{I}}^{m_2}_i, \hat{\mathcal{I}}^{m_3}_i), (\bar{\mathcal{I}}^{m_1}_i, \bar{\mathcal{I}}^{m_2}_i, \bar{\mathcal{I}}^{m_3}_i)\big) + \lambda \cdot \mathcal{L}_{cls}(\hat{L}_{t_i}, L_{t_i})\Big)
\]
where $\mathcal{L}_{gen}(\cdot, \cdot)$ is the generation cross-entropy loss, while $\mathcal{L}_{cls}(\cdot, \cdot)$ is the classification cross-entropy loss based on the $cls$, and $\lambda\in\mathbb{R}$ is a hyperparameter. 
Some patients do not have a previous scan, i.e.  $\bar{\mathcal{I}}^{x_{(t_{j-1})}}=\emptyset$. We replace all missing previous scans with a set of learnable padding tokens to ensure a consistent model input size during training. We follow a similar approach to Eq. \eqref{eq:tranformer_func} during inference without providing the modality we wish to generate, i.e., the input of $\mathcal{M}_\theta$ is a patient's available modalities.

\noindent \textbf{Causal Attention:}
Inputting multiple modalities raises computational complexity challenges associated with the standard attention mechanism, which exhibits $\mathcal{O}(N^2)$ where $N$ being the number of tokens.
As such, we employ the Performer architecture along with the causal attention mechanism \cite{performer} that reduces the complexity down to $\mathcal{O}(N)$. This is achieved by approximating the Softmax function, enabling the multiplication of the key and value matrices to be carried out before the result is multiplied by the query matrix. Our proposed model $\mathcal{M}_\theta$ generates tokens in an \textit{auto-regressive} manner, inserting each predicted token into the next step to generate the next token. To preserve this causality and allow multiprocessing, we implement unidirectional FAVOR$+$ \cite{performer}. The causal constraint prevents a token from attending to tokens positioned after it in the sequence, thereby ensuring no information leakage from future tokens. Following our Performer's linear projection for the different inputs, we inject sinusoidal positional embedding to the text embedding, as well as axial positional embedding \cite{ho2019axial} to the previous and current image embeddings. We also inject sinusoidal positional embedding into the full input embeddings before passing it to our model.

\section{Datasets and Experiments}
\textbf{Datasets:}
\label{datasetsection}
We extensively test our method, TiBiX, on a selected subset of MIMIC-CXR \cite{Johnson2019Dec}. We also introduce MIMIC-T, a new dataset derived from MIMIC-CXR, containing longitudinal patient data. Our preprocessing involves discarding studies lacking both impressions and findings sections and retaining only AP frontal-view scans. This yields approximately $122k$, $1k$, $2k$ unique studies from $32521$, $255$, $266$ patients in train, validate, and test splits, respectively. To increase the number of training samples, we utilize prior scans by considering every consecutive pair for the same subject as a new sample. Moreover, we also consider single-study cases simulating real-world scenarios. We prioritize examining previous scans to diagnose diseases and assess prognosis. The distribution of patients with one or two studies is outlined in Section \ref{sec:app_sec_a} in the Appendix.\\
\textbf{Implementation Details:}
\label{reportpre}
We extract impressions and findings from reports and preprocess them by converting text to lowercase, removing punctuation, and digits. We use the CXR-BERT \cite{cxrbert} tokenizer with 30,522 unique tokens, including special tokens. AP frontal-view images are resized to 512x512. We employ a pretrained VQ-GAN from \cite{unixgen}, trained on MIMIC-CXR with dz=256 and a codebook size of 1024, encoding 512x512 images into 32x32=1024 tokens. Our transformer-based model uses the same configuration of Performer \cite{performer}. It handles 1029 image tokens and 30,522 text tokens \cite{vqgan,performer}. The model can process two images if the previous one is available. Training lasts for 150 epochs with a batch size of 10, AdamW optimizer, learning rate of 2e-4, weight decay of 1e-2, and cosine decay schedule. Report and image generation use Top-p sampling (p=0.9, temperature=0.7). We ensemble five model weights and select the image with the highest pixel entropy to mitigate stochastic blank image generation.

\section{Results and Discussion}

\begin{table*}[t]
\caption{Comparison to the previous methods in report generation. (\#) - Subset of MIMIC-T with $n$ studies where \( n \in [1, 2] \quad \text{and} \quad n \in \mathbb{N}\).}
\label{table4}
\centering
\resizebox{\textwidth}{!}{
\begin{tabular}{p{0.21\linewidth} |c|cccccc|ccc}
\hline
\textbf{Methods} &
  \textbf{Set (\#)} &
  \multicolumn{6}{c|}{\textbf{NLG metric}} &
  \multicolumn{3}{c}{\textbf{Clinical Efficacy(CE)}} \\ \hline
 &
  \multicolumn{1}{l|}{} &
  \textbf{BLEU-1} &
  \textbf{BLEU-2} &
  \textbf{BLEU-3} &
  \textbf{BLEU-4} &
  \textbf{ROUGE} &
  \textbf{METEOR} &
  \textbf{Precision} &
  \textbf{Recall} &
  \textbf{F1} \\ \hline
{CNN-RNN \cite{cnnrnn}}             & CR$|$CX (2)     & 0.190 & 0.105 & 0.066 & 0.045 & 0.200 & 0.096 & 0.303 & 0.221         & 0.230          \\ 
{AdaAtt \cite{knowingwheretolook}}             & CR$|$CX (2)      & 0.206 & 0.115 & 0.072 & 0.049 & 0.205 & 0.102 & 0.310 & 0.230          & 0.234          \\ 
{Transformer \cite{transformer}}         & CR$|$CX (2)      & 0.194 & 0.109 & 0.068 & 0.047 & 0.200 & 0.100 & 0.320 & 0.228          & 0.241          \\ 
{M2Transformer\cite{M2v1}} & CR$|$CX (2)      & 0.193 & 0.108 & 0.068 & 0.047 & 0.200 & 0.099 & \underline{0.323} & \underline{0.248}          & \underline{0.253}          \\ 
{R2GenCMN \cite{r2gencmn}} &
  CR$|$CX (2) &
  \textbf{0.348} &
  \textbf{0.240} &
  \underline{0.182} &
  \underline{0.146} &
  0.322 &
  \underline{0.148} &
  {0.321} &
  {0.216} &
  {0.244} \\ 
{XproNet \cite{xpronet}}            & 
    CR$|$CX (2)      & 
    \underline{0.324} & 
    0.225 & 
    0.172 & 
    0.139 & 
    \underline{0.323} &
    0.141 &
    0.315 &
    0.211 & 
    0.240 \\ \hline
  & CR$|$CX (1)     & 0.234 & 0.156 & 0.116 & 0.091 & 0.272 & 0.119 & 0.340          & 0.190 & 0.230 \\    
  &          CR$|$CX (2)      &
    0.284 & 
    0.184 & 
    0.136 & 
    0.105 & 
    0.267 & 
    0.127 & 
    \textbf{0.340} & 
    0.220 & 
    \underline{0.250} \\
\multirow{-2}{*} {\textbf{TiBiX(Ours)}} &
  CR$|$(PX,CX) (2) & 
    \underline{0.324} & 
    \underline{0.234} &
    \textbf{0.185} &
    \textbf{0.157} &
    \textbf{0.331} &
    \textbf{0.162} &
    0.300 &
    \underline{0.224} &
    \underline{0.25} \\ \hline
\end{tabular}}
\end{table*}

\begin{table*}[t]
\caption{Comparison to state-of-the-art (SOTA) in regard to image  generation task. (\#) - Subset of MIMIC-T with $n$ studies where \( n \in [1, 2] \quad \text{and} \quad n \in \mathbb{N}\).}
\label{table5}

\centering
\resizebox{\textwidth}{!}{
\begin{tabular}{c|c|cccc|ccccc} 

\hline

   & 
   \multicolumn{1}{c|}{} &
  \multicolumn{4}{c|}{\textbf{Image Generation metrics}} &
  \multicolumn{5}{c}{\textbf{14 diagnosis classification}} \\ \cline{3-11} 
\multirow{-2}{*}{\textbf{Method}} &

  \multirow{-2}{*}{\textbf{Set (\#)}} &
  \textbf{FID} &
  \textbf{SSIM} &
  \textbf{MSSIM} &
  \textbf{IS} &
  \textbf{Precision} &
  \textbf{Recall} &
  \textbf{AUC} &
  \textbf{Accuracy} &
  \textbf{F1} \\ \hline

\multirow{-2}{*}{} &

  GT (2) &
  - &
  - &
  - &
  - &
  0.75 &
  0.73 &
  0.79 &
  0.75 &
  0.60 \\

{Roentgen \cite{roentgen}} &

  CX$|$CR (2) &
  81.02 &
  0.51 &
  0.398 &
  \underline{1.96\textpm0.08} &
  \textbf{0.64} &
  \underline{0.64} &
  \textbf{0.84} &
  \textbf{0.77} &
  \textbf{0.60} \\ 

{Cascaded Diffusion\cite{weber2023cascaded}} &
 
  CX$|$CR (2) &
  \textbf{28.78} &
  \textbf{0.53} &
  \underline{0.428} &
  \textbf{2.05\textpm0.06} &
  0.48 &
  \textbf{0.82} &
  \underline{0.75} &
  0.66 &
  \underline{0.57} 
\\ \hline
 &
CX$|$CR (1) & 80.48      &  0.50         & 0.442          & 1.70\textpm0.15   & 0.68              & 0.28           & 0.69        & 0.70             & 0.42       \\ 

 &
   CX$|$CR (2) &
  50.40 &
  0.50 &
  0.438 &
  1.76\textpm0.10 &
  \underline{0.61} &
  0.35 &
   0.69 &
    \underline{0.74} &
  0.41 \\
 \multirow{-2}{*} {\textbf{TiBiX(Ours)}} &

   CX$|$(CR,PX) (2) &
  \underline{33.85} &
  \underline{0.52} &
  \textbf{0.472} &
  1.82\textpm0.08 &
  0.56 &
  0.55 &
  \underline{0.75} &
  \textbf{0.77} &
  0.53 \\ \hline
\end{tabular}}
\end{table*}

\subsection{SOTA Performance Comparison}
To demonstrate the effectiveness of our approach, we compare with several SOTA methods in the domain of report and CXR image generation. In all the Tables and Figures below we use the following notations: GT - Ground Truth, CR - Current Report, CX - Current CXR, PX - Previous CXR.(CX$|$CR) denotes the current CXR generation given the current report, (CX$|$(CR,PX)) denotes the current CXR generation given the current report and previous CXR.
\newline
\textbf{Report Generation:} We compare our results on the report generation task with several SOTA image-captioning works such as the CNN-RNN\cite{cnnrnn}, Transformer \cite{transformer}, AdaAtt \cite{knowingwheretolook}, M2Transfomer \cite{M2v1} in the natural domain as well as the SOTA works in the medical domain including R2GenCMN \cite{r2gencmn} and XproNet \cite{xpronet}.  For \cite{showattendtell,cnnrnn,transformer,M2v1} methods, we rerun them on our datasets, following the official implementation of \cite{M2v1}. We adhere to the same preprocessing steps as in our approach, while keeping the other hyperparameters consistent with those specified in their codebase. Table \ref{table4} shows that compared to other works, our method achieves SOTA results on most of the NLG metrics and also achieves an on-par performance on the classification scores.
\newline
\textbf{CXR Image Generation:} Few works in the medical domain have focused on the generation of Chest X-ray images. We compare our results with the recent top works in the field of X-ray generation, which have used Diffusion Models(DM) \cite{roentgen,weber2023cascaded}. We have implemented these pretrained models on the subset of the test set having 2 studies, and the results are shown in Table \ref{table5}. RoentGen is a medical domain-adapted latent diffusion model based on the Stable Diffusion pipeline whereas Cascaded Diffusion\cite{weber2023cascaded} improves upon this work, by having cascaded models.  The first model in \cite{weber2023cascaded} generates a lower resolution image and the second one generates a higher resolution image, enhancing the first DM's output. It is noteworthy that \cite{weber2023cascaded} is trained on a huge dataset Massive Chest X-ray Dataset (MaCheX) around 3.5x the size of MIMIC-T. Our results outperform Roentgen's and are comparable to the Cascaded Diffusion approach in terms of image generation metrics. However, we have found that Roentgen outperforms all other methods regarding classification scores. It is counterintuitive that RoentGen's performance in image generation is poor, whereas it performs exceptionally well when the generated images are passed through a classifier. This could be due to the pathology being easily detected by the model, which is generally not the case in real life.

It is worth noting that all the works we compare are specifically designed for a single task, either reporting or image generation. In contrast, our framework excels in bidirectional generation while achieving state-of-the-art performance across all tasks.


\subsection{Self-Comparison}
We analyse different subsets of the MIMIC-T dataset to understand the effectiveness and importance of including a prior CXR scan for generation. In Table \ref{table4}, we evaluate the quality and clinical efficacy (CE) of the report generation task on 2 subsets of the test set : (1) patients with 1 study and (2) patients with 2 studies. Within the second case, we analyse CR $|$ CX as well as the case of CR $|$ (PX,CX). We observe that when the prior scan is given along with the current report, the BLEU-4 increases by around 1.5x times. Furthermore, we observe significant improvement across all the metrics including Clinical Efficacy. These results, are analogous to the fact that radiologists when provided with multiple scans, can produce more comprehensive and accurate diagnostic reports for various diseases. Similarly in Table \ref{table5}, we compare the results on the different subsets of MIMIC-T, for the image generation task using FID, SSIM, MSSIM and IS. Furthermore, we also run a classifier (MIMIC-CXR pretrained DenseNet121) on the images and get the classification scores. Given both the prior CXR and CR, the generation quality improves significantly. Furthermore, it achieves the best results in all the evaluation metrics including the classification results. We also provide some ablation studies in Section \ref{sec:app_sec_c} in the Appendix.

\begin{figure*}[htb]
    \centering
    \includegraphics[width=0.93\textwidth]{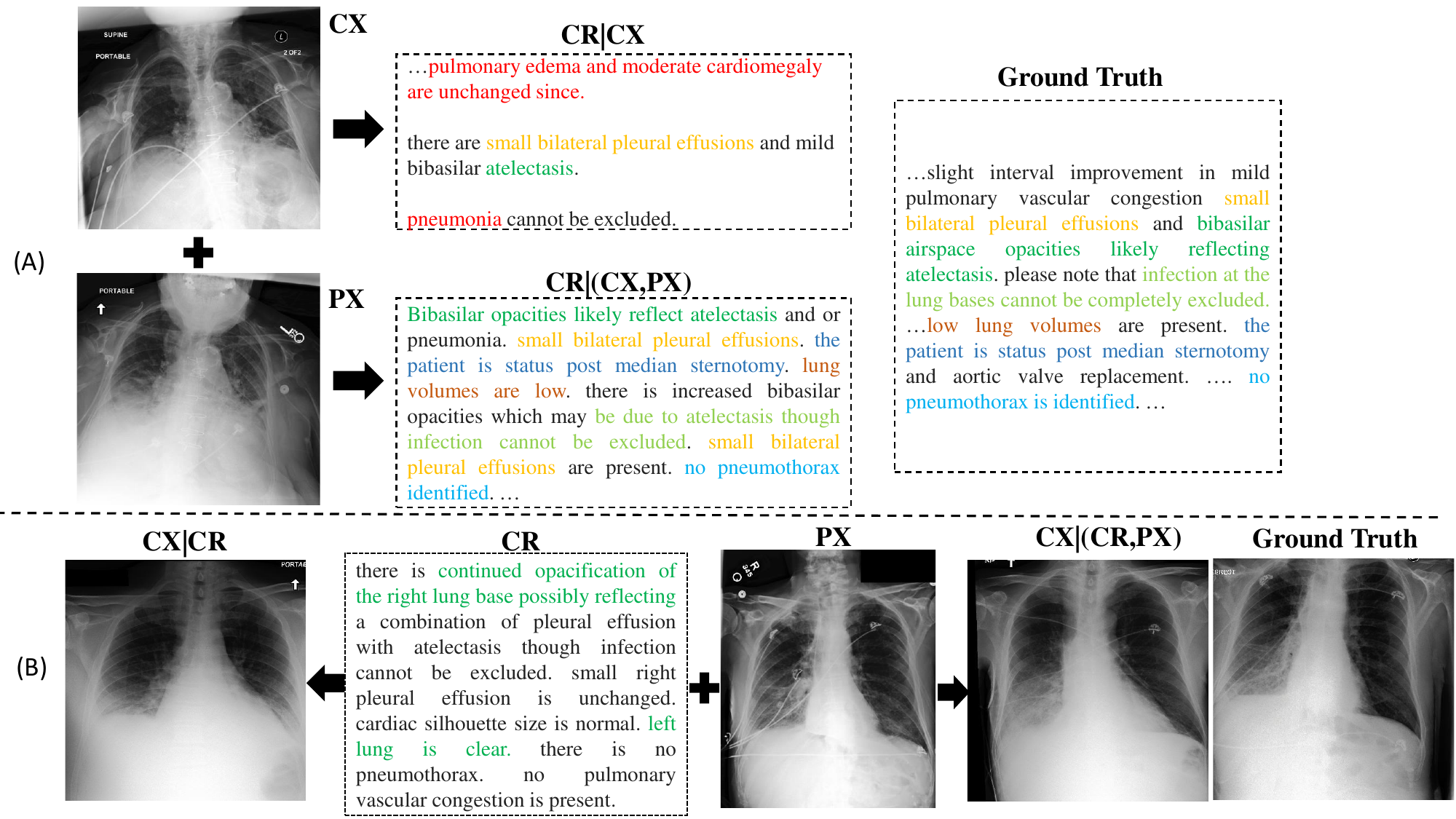} 
    \caption{(A) Report Generation Task: The first case describes an example of (1) current report generation (CR$|$CX) (2) current report generation (CR$|$(PX,CX)). (B) CXR Image Generation task: The left side of the report is (CX$|$CR) and the on the right side is (CX$|$(CR,PX)). The GT is present for comparison.}
    \label{fig:qualitative}
\end{figure*}


\subsection{Qualitative Analysis}
The benefits of considering the prior scan when generating a report or a scan are demonstrated in Fig. \ref{fig:qualitative}. Fig. \ref{fig:qualitative}-A illustrates that in the case where the model does not consider the prior scan it produces a report with false findings that are not in the ground truth (e.g. the report describes cardiomegaly when it's not there). Furthermore, it hallucinates and includes the word "unchanged" even though the previous scan is not provided. On the other hand, when the model considers the previous scan it generates a report with findings more aligned with the ground truth (highlighted in different colors). Fig. \ref{fig:qualitative}-B illustrates that when the model considers the previous scan, it generates a scan with an anatomical structure more similar to the ground truth. The resultant scan is also closer to the ground truth in terms of portraying the pathologies. For example, ``the opacification of the right lung base as opposed to the left lung being clear is more accurate when the models take into account the previous scan''. When the models do not have the prior scan, the resultant image deviates from the ground truth more. We also provide extensive qualitative analysis comparing previous methods in Section \ref{sec:app_sec_b} in the Appendix.

\section{Conclusion}
In this work, we introduce a novel approach to enhance image and report generation by integrating temporal data from previous CXR scans. Our framework includes a transformer-based model with causal attention, facilitating bidirectional CXR-to-report generation while considering the time difference between consecutive scans. Experimental results on the MIMIC-T dataset, a temporal subset of MIMIC-CXR, demonstrate promising performance. Qualitative analysis and ablation studies justify the inclusion of prior scans and highlight the importance of each architectural component. While our work lays a foundational framework, future enhancements could explore incorporating prior reports and domain expertise through knowledge graphs. Furthermore, we aim to create a metric capable of capturing the temporal aspect, addressing the inadequacies of current metrics in this regard.

\bibliographystyle{splncs04}
\bibliography{references}

\appendix

\newpage

\appendix
\section*{Appendix}
\vspace{-5pt}

\renewcommand{\thesection}{A\arabic{section}}
\setcounter{section}{0}
\setcounter{figure}{0}
\setcounter{table}{0}

\section{Dataset split}
\vspace{-20pt}
\label{sec:app_sec_a}

\begin{table}[h!]
\centering
{
\caption{Number of patients and studies in the MIMIC-T dataset.}
\label{datasettable}
\begin{tabular}{llll}
\hline
\textbf{\#Number of patients} & \textbf{Train} & \textbf{Validation} & \textbf{Test} \\ \hline
with 1 study                  & 35498            & 282                 & 299           \\
with 2 studies                & 112393           & 1088                & 2066          \\ \hline
Total studies(scans)          & 260284          & 2298                & 4431          \\ \hline
\end{tabular}
}
\end{table}

\vspace{-2.em}

\renewcommand{\thesection}{B\arabic{section}}
\setcounter{section}{0}
\setcounter{figure}{0}
\setcounter{table}{0}

\section{Qualitative Comparison of X-ray and report generation tasks against various methods} 
\label{sec:app_sec_b}
\vspace{-2.em}

\begin{table}[h!]
\centering

{
    \caption{Comparison for the Report Generation Task with other Report Generation methods such as R2Gen, R2GenCMN, XProNet. CX - Current CXR, PX- Previous CXR, CR - Current Report. }
    \includegraphics[width=0.84\textwidth]{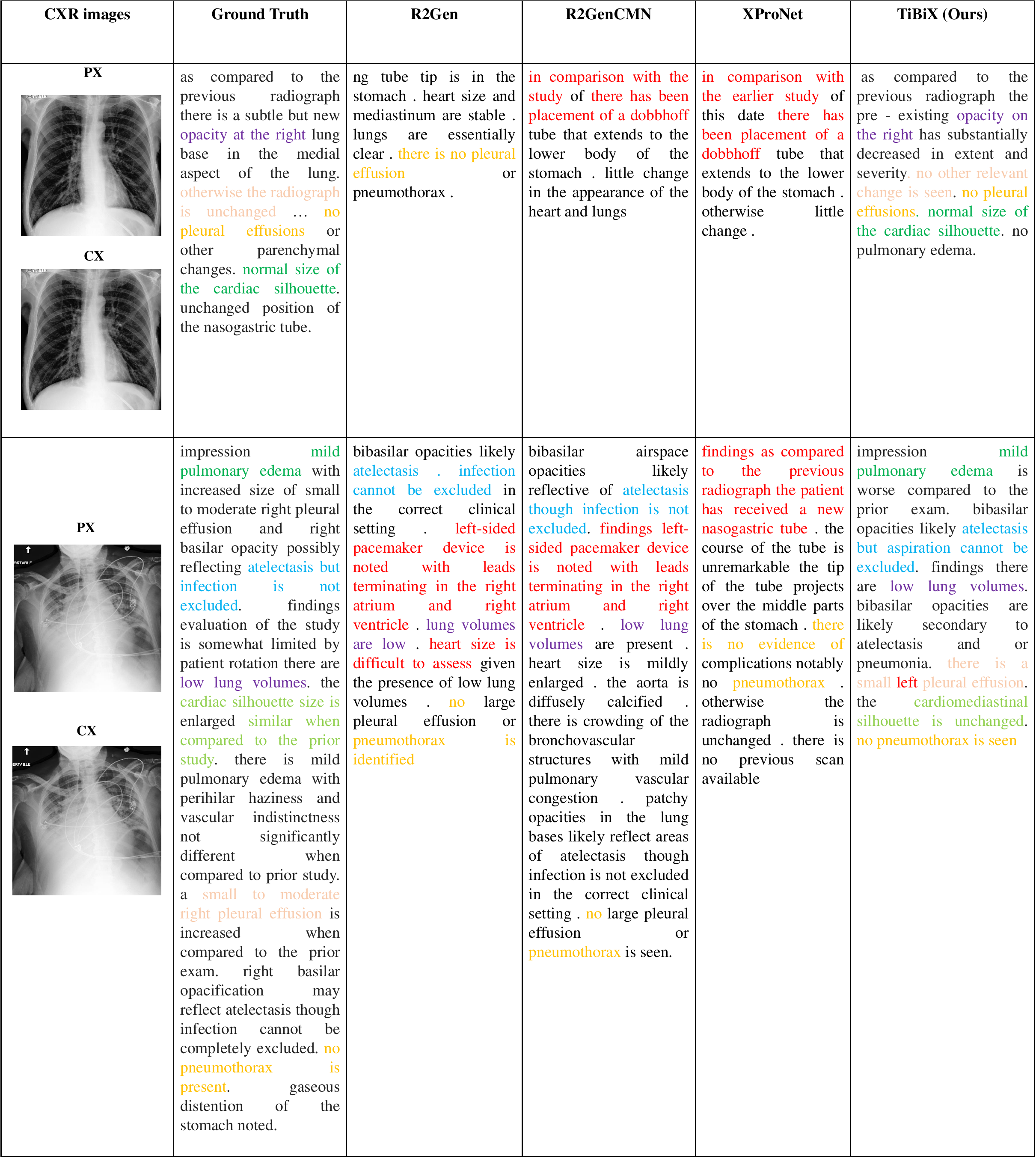} 
    \label{fig:report_generation}
}
\end{table}

\begin{table}[h!]
\centering
{
    \caption{Comparison for the Image Generation Task with other Image Generation methods -  RoentGen and Cascaded Diffusion. CX - Current CXR, PX- Previous CXR, CR - Current Report. }
    \includegraphics[width=\textwidth]{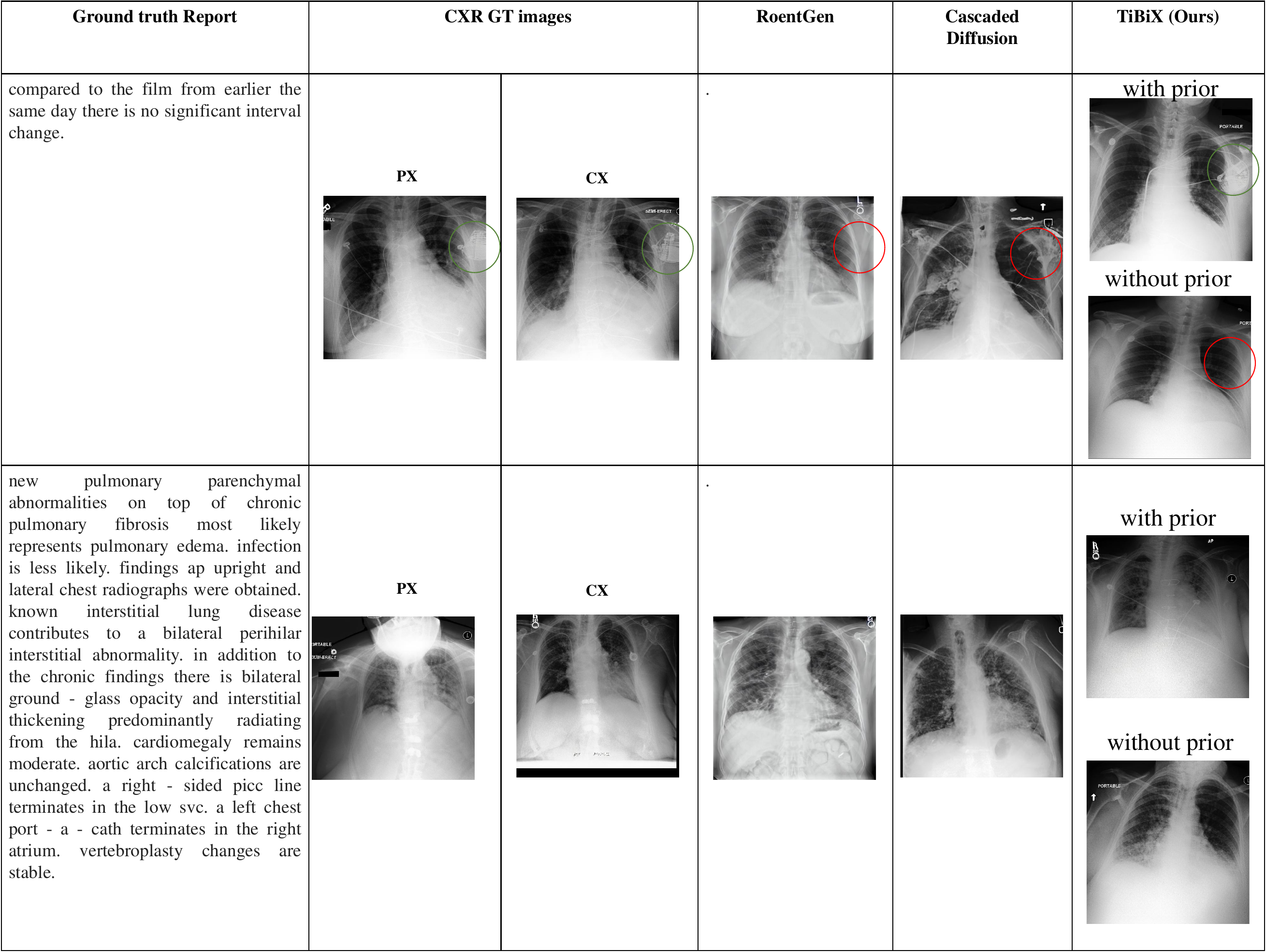} 
    \label{fig:image_generation}
}
\end{table}

\renewcommand{\thesection}{C\arabic{section}}
\setcounter{section}{0}
\setcounter{figure}{0}
\setcounter{table}{0}
\section{Ablation Study}
\label{sec:app_sec_c}

\vspace{-1.em}

\begin{table}[h!]
\centering
{
\caption{Ablation Study on the inclusion of different components of TiBiX. BASE-Baseline model, CLS - Class token, TT - Temporal token}
\label{table3}
\resizebox{\textwidth}{!}{

\begin{tabular}{c|cccc|ccccc}
\hline
\textbf{Method} &
  \multicolumn{4}{c|}{\textbf{Image Generation metrics}} &
  \multicolumn{5}{c}{\textbf{14 diagnosis classification}} \\ \hline
 &
  \textbf{FID} &
  \textbf{SSIM} &
  \textbf{MSSIM} &
  \textbf{IS} &
  \textbf{Precision} &
  \textbf{Recall} &
  \textbf{AUC} &
  \textbf{Accuracy} &
  \textbf{F1} \\ \hline
Base (CXRBERT + VQGAN)  & 88.51 & 0.383 & 0.396 & 2.42\textpm0.09  & 0.49 & 0.54 & \textbf{0.69} & \underline{0.71} & 0.48 \\
Base + CLS & \underline{81.83} & \underline{0.396} & \textbf{0.396} & \underline{2.64\textpm0.07} & \underline{0.50} & 0.54 & \underline{0.68} & \textbf{0.72} & \textbf{0.50} \\

Base(Bi) + CLS + TT        & 89.58 & 0.376 & \underline{0.387} & 2.50\textpm0.11 & \textbf{0.51} & \textbf{0.54} & \textbf{0.69} & \textbf{0.72} & \underline{0.49} \\

Unidirectional Image (+ CLS + TT) & \textbf{70.11} & \textbf{0.415} & 0.371 & \textbf{2.96\textpm0.18}& 0.44 & 0.45 & 0.60 & 0.68 & 0.42 \\ \hline

\begin{tabular}[c]{@{}c@{}} \textbf{Base(Bi) + CLS + TT (w ensemble)} \end{tabular} &
  \textbf{33.85} &
  \textbf{0.515} &
  \textbf{0.472} &
  1.82\textpm0.08 &
  \textbf{0.56} &
  \textbf{0.55} &
  \textbf{0.75} &
  \textbf{0.77} &
  \textbf{0.53} \\ \hline
\end{tabular}}
}
\vspace{1em}
{
\caption{Comparison of unidirectional and bidirectional methods for the report generation task}
\label{table1}%
\begin{tabular}{c|cc|ccc}
\hline
\textbf{Method}     & \multicolumn{2}{c|}{\textbf{NLG metrics}} & \multicolumn{3}{c}{\textbf{CE}}                \\ \hline
                      & \textbf{BLEU-3} & \textbf{BLEU-4} & \textbf{P} & \textbf{R} & \textbf{F1} \\ \hline
Unidirectional Report & 0.172           & 0.146           & 0.292      & 0.200       & 0.222       \\
Bidirectional(Ours) & \textbf{0.185}      & \textbf{0.157}      & \textbf{0.300} & \textbf{0.224} & \textbf{0.250} \\ \hline
\end{tabular}
}
\end{table}

\end{document}